\documentclass[a4paper,fleqn]{cas-dc}
\usepackage{booktabs}

\usepackage[numbers]{natbib}
\def\tsc#1{\csdef{#1}{\textsc{\lowercase{#1}}\xspace}}
\tsc{CNN}
\tsc{Transformers}
\tsc{EP}
\tsc{PMS}
\tsc{BEC}
\tsc{DE}
\usepackage{graphicx}

\begin{document}

\shortauthors{Peng Zhao et~al.}
\let\printorcid\relax
\title [mode = title]{EMDS-6: Environmental Microorganism Image
Dataset Sixth Version for Image Denoising, Segmentation, Feature Extraction, Classification and Detection Methods Evaluation}

\author[a]{Peng Zhao}

\author[a]{Chen Li}
\cormark[1]
\ead{lichen201096@hotmail.com}

\author[a, f]{Md Mamunur Rahaman}

\author[a]{Hao Xu}

\author[a]{Pingli Ma}

\author[a]{Hechen Yang}

\author[d]{Hongzan Sun}

\author[b]{Tao Jiang}[style=chinese]
\cormark[2]
\ead{jiang@cuit.edu.cn}

\author[e]{Ning Xu}

\author[c]{Marcin Grzegorzek}

\address[a]{Microscopic Image and Medical Image Analysis Group, MBIE College, Northeastern University, 110169, Shenyang, PR China}

\address[b]{School of Control Engineering, Chengdu 
University of Information Technology, Chengdu 610225, China}

\address[c]{University of Lübeck, Germany}

\address[d]{Department of Radiology, Shengjing Hospital, China Medical University, Shenyang, 110122, China}
\cortext[cor1]{Corresponding author}

\address[e]{School of Arts and Design, Liaoning Shihua University, Fushun 113001, China}

\address[f]{School of Computer Science and Engineering, University of New South Wales, Sydney, Australia}

\begin{abstract}
\  
Environmental microorganisms (EMs) are ubiquitous around us and have an important impact on the survival and development of human society. However, the high standards and strict requirements for the preparation of environmental microorganism (EM) data have led to the insufficient of existing related datasets, not to mention the datasets with ground
truth (GT) images. This problem seriously affects the progress of related experiments. Therefore, This study develops the \emph{Environmental Microorganism Dataset Sixth Version} (EMDS-6), which contains 21 types of EMs. Each type of EM contains 40 original and 40 GT images, in total 1680 EM images. In this study, in order to test the effectiveness of EMDS-6. We choose the classic algorithms of image processing methods such as image denoising, image segmentation and target detection. The experimental result shows that EMDS-6 can be used to evaluate the performance of image denoising, image segmentation, image feature extraction, image classification, and object detection methods. EMDS-6 is available at the \url{https://figshare.com/articles/dataset/EMDS6/17125025/1}.

\end{abstract}


\maketitle

\section{Introduction}

\subsection{Environmental Microorganisms}

\emph{Environmental Microorganisms} (EMs) usually refer to tiny living that exists in nature and are invisible to the naked eye and can only be seen with the help of a microscope. Although EMs are tiny, they significantly impacts human survival~\cite{madigan_1997_brock}. Some beneficial bacteria can be used to produce fermented foods such as cheese and bread from a beneficial perspective. Meanwhile, Some beneficial EMs can degrade plastics, treat sulfur-containing waste gas in industrial, and improve the soil. From a harmful point of view, EMs cause food spoilage, reduce crop production and are also one of the chief culprits leading to the epidemic of infectious diseases. To make better use of the advantages of environmental microorganisms and prevent their harm, a large number of scientific researchers have joined the research of EMs. The image analysis of EM is the foundation of all this.

EMs are tiny in size, usually between 0.1 and 100 microns. This poses certain difficulties for the detection and identification of EMs. Traditional "morphological methods" require researchers to look directly under a microscope~\cite{madsen_2015_environmental}. Then, the results are presented according to the shape characteristics. This traditional method requires more labor costs and time costs. Therefore, using computer-assisted feature extraction and analysis of EM images can enable researchers to use their least professional knowledge with minimum time to make the most accurate decisions.

\subsection{EM Image Processing  and Analysis}

Image analysis is a combination of mathematical models and image processing technology to analyze and extract certain intelligence information. Image processing refers to the use of computers to analyze images. Common processing includes image denoising, image segmentation and feature extraction. Image noise refers to various factors in the image that hinder people from accepting its information. Image noise is generally generated during image acquisition, transmission and compression~\cite{pitas_2000_digital}. The aim of image denoising is to recover the original image from the noisy image~\cite{buades_2005_review}. Image segmentation is a critical step of image processing to analyze an image. In the segmentation, we divide an image into several regions with unique properties and extract regions of interest~\cite{minaee_2021_image}. Feature extraction refers to obtaining important information from images such as values or vectors~\cite{zebari_2020_comprehensive}. Moreover, these characteristics can be distinguished from other types of objects. Using these features, we can classify images. Meanwhile, the features of an image are the basis of object detection. Object detection uses algorithms to generate object candidate frames, that is, object positions. Then, classify and regress the candidate frames.

\subsection{The Contribution of Environmental Microorganism Image Dataset Sixth Version (EMDS-6):}

Sample collections of the EMs are usually performed outdoors. When transporting or moving samples to the laboratory for observation, drastic changes in the environment and temperature affect the quality of EM samples. At the same time, if the researcher observes EMs under a traditional optical microscope, it is very prone to subjective errors due to continuous and long-term visual processing. Therefore, the collection of environmental microorganism image datasets is challenging~\cite{kosov_2018_environmental}. Most of the existing environmental microorganism image datasets are not publicly available. This has a great impact on the progress of related scientific research. For this reason, we have created the \emph{Environmental Microorganism Image Dataset Sixth Version} (EMDS-6) and made it publicly available to assist related scientific researchers. Compared with other environmental microorganism image datasets, EMDS-6 has many advantages. The dataset contains a variety of microorganisms and provides possibilities for multi-classification of EM images. In addition, each image of EMDS-6 has a corresponding ground truth (GT) image. GT images can be used for performance evaluation of image segmentation and target detection. However, the GT image production process is extremely complicated and consumes enormous time and human resources. Therefore, many environmental microorganism image dataset does not have GT images. However, our proposed dataset has GT images. In our experiments, EMDS-6 can provide robust data support in tasks such as denoising, image segmentation, feature extraction, image classification and object detection. Therefore, the main contribution of the EMDS-6 dataset is to provide data support for image analysis and image processing related research and promote the development of EMs related experiments and research.

\section{Materials and Methods}

\subsection{EMDS-6 Dataset}

\begin{figure*}
	\centering
	\includegraphics[scale=0.51]{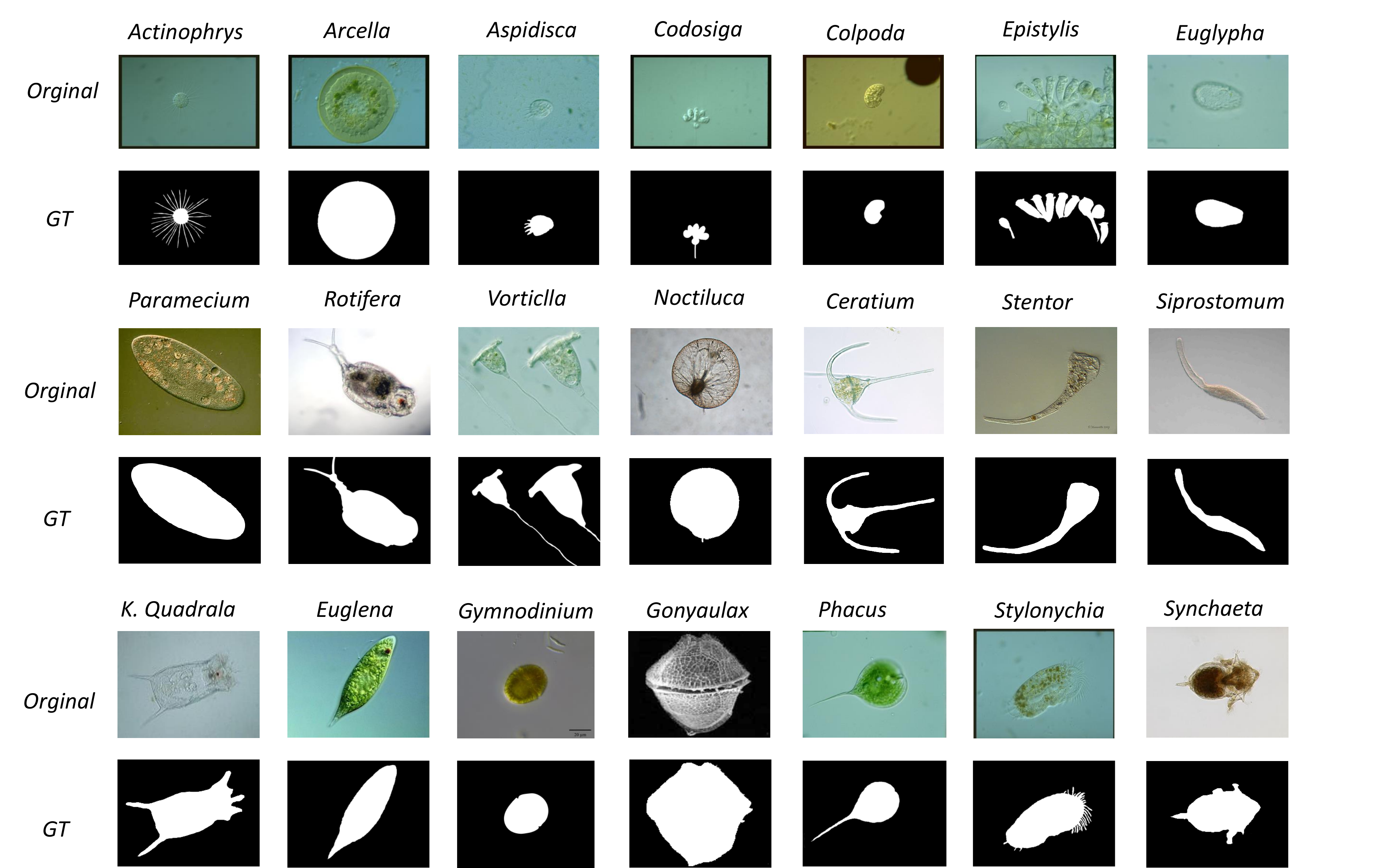}
	
	\caption{An example of EMDS-6, including original images and GT images.}
\label{FIG:1}
\end{figure*}

There are 1680 images in the EMDS-6 dataset, including 21 classes of original EM images with 40 images per class, resulting in a total of 840 original images, and each original image is followed by a GT image for a total of 840. Table. \ref{tbl1} shows the details of the EMDS-6 dataset. Figure. \ref{FIG:1} shows some examples of the original images and GT images in EMDS-6. EMDS-6 is freely published for non-commercial purpose at: \url{https://figshare.com/articles/dataset/EMDS6/17125025/1}.

\begin{table*}[htbp]

\small
	\caption{Basic information of EMDS-6 dataset, including Number of original images (NoOI), Number of GT images (NoGT).}
\renewcommand\arraystretch{1.8} 
\setlength{\tabcolsep}{6mm}{
\begin{tabular}{|c|c|c|c|c|c|c|c|}
\hline
Class       & NoOI & NoGT & Class       & NoOI & NoGT \\ \hline
Actinophrys & 40   & 40   & Ceratium    & 40   & 40   \\ \hline
Arcella     & 40   & 40   & Stentor     & 40   & 40   \\ \hline
Aspidisca   & 40   & 40   & Siprostomum & 40   & 40   \\ \hline
Codosiga    & 40   & 40   & K. Quadrala & 40   & 40   \\ \hline
Colpoda     & 40   & 40   & Euglena     & 40   & 40   \\ \hline
Epistylis   & 40   & 40   & Gymnodinium & 40   & 40   \\ \hline
Euglypha    & 40   & 40   & Gonyaulax   & 40   & 40   \\ \hline
Paramecium  & 40   & 40   & Phacus      & 40   & 40   \\ \hline
Rotifera    & 40   & 40   & Stylonychia & 40   & 40   \\ \hline
Vorticlla   & 40   & 40   & Synchaeta   & 40   & 40   \\ \hline
Noctiluca   & 40   & 40   & -           & -    & -    \\ \hline
Total       & 840  & 840  & Total       & 840  & 840  \\ \hline
\end{tabular}}
\label{tbl1}
\end{table*}

The collection process of EMDS-6 images starts from 2012 till 2020. The following people have made a significant contribution in producing the EMDS-6 dataset: Prof. Beihai Zhou and Dr Fangshu Ma from the University of Science and Technology Beijing, China; Prof. Dr.-Ing. Chen Li and M.E. HaoXu from Northeastern University, China; Prof. Yanling Zou from Heidelberg University, Germany. The GT images of the EMDS-6 dataset are produced by Prof. Dr.-Ing Chen Li, M.E. Bolin Lu, M.E. Xuemin Zhu and B.E. Huaqian Yuan from Northeastern University, China. The GT image labelling rules are as follows: the area where the microorganism is located is marked as white as foreground, and the rest is marked as black as the background.

\subsection{Experimental Method and Setup}

To better demonstrate the functions of EMDS-6, we carry out noise addition and denoising experiments, image segmentation experiments, image feature extraction experiments, image classification experiments and object detection experiments. The experimental methods and data settings are shown below. Moreover, we select different critical indexes to evaluate each experimental result in this section.

\subsubsection{Noise Addition and Denoising Method}
In digital image processing, the quality of an image to be recognized is often affected by external conditions, such as input equipment and the environment. Noise generated by external environmental influences largely affects image processing and analysis (e.g., image edge detection, classification, and segmentation). Therefore, image denoising is the key step of image preprocessing.

\begin{figure*}
	\centering
	\includegraphics[scale=0.53]{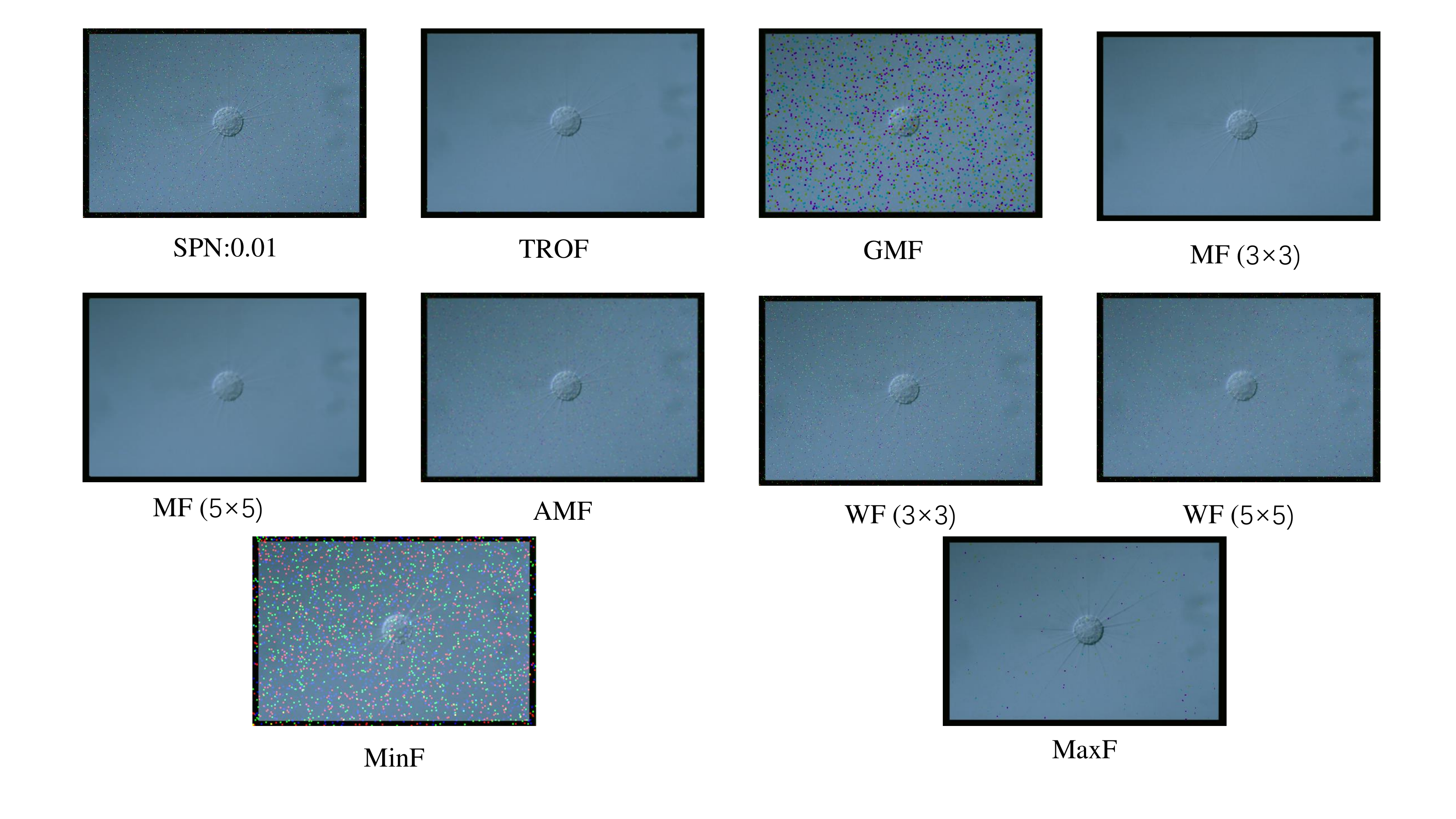}
	
	\caption{Examples of using different filters to filter salt and pepper noise.}
	\label{FIG:2}
\end{figure*}

In this study, we have used four types of noise, Poisson noise, multiplicative noise, Gaussian noise and pretzel noise. By adjusting the mean, variance and density of different kinds of noise, a total of 13 specific noises are generated. They are multiplicative noise with a variance of 0.2 and 0.04 (marked as MN:0.2 and MN: 0.04 in the table), salt and pepper noise with a density of 0.01 and 0.03 (SPN:0.01, SPN:0.03), pepper noise (PpN), salt noise (SN), Brightness Gaussian noise (BGN), Positional Gaussian noise (PGN), Gaussian noise with a variance of 0.01 and a mean of 0 (GN 0.01-0), Gaussian noise with a variance of 0.01 and a mean of 0.5 (GN 0.01-0.5), Gaussian noise with a variance of 0.03 and a mean of 0 (GN 0.03-0), Gaussian noise with a variance of 0.03 and a mean of 0.5 (GN 0.03-0.5), and Poisson noise (PN). There are 9 kinds of filters at the same time, namely Two-Dimensional Rank Order Filter (TROF), 3×3 Wiener Filter (WF ($3\times3$)), 5×5 Wiener Filter (WF ( $5\times5$)), 3×3 Window Mean Filter (MF ($3\times3$)), Mean Filter with 5×5 Window  (MF ($5\times5$)). Minimum Filtering (MinF), Maximum Filtering (MaxF), Geometric Mean Filtering (GMF), Arithmetic Mean Filtering (AMF). In the experiment, 13 kinds of noise are added to the EMDS-6 dataset image, and then 9 kinds of filters are used for filtering. The result of adding noise into the image and filtering is shown in Figure.~\ref{FIG:2}. 

\subsubsection{Image Segmentation Methods}

This paper designs the following experiment to prove that EMDS-6 can be used to test different image segmentation methods. Six classic segmentation methods are used in the experiment: $k$-means \cite{burney_2014_k-means}, Markov Random Field (MRF) \cite{kato_2012_markov}, Otsu Thresholding \cite{otsu_1979_threshold}, Region Growing (REG) \cite{adams_1994_seeded},  Region Split and Merge Algorithm (RSMA) \cite{chen_1991_split} and Watershed Segmentation \cite{levner_2007_classification} and one deep learning-based segmentation method, Recurrent Residual CNN-based U-Net (U-Net)~\cite{alom_2019_recurrent} are used in this experiment. While using U-Net for segmentation, the learning rate of the network is 0.001 and the batch size is 1. In the $k$-means algorithm, the value of $k$ is set to 3, the initial center is chosen randomly, and the iterations are stopped when the number of iterations exceeds the maximum number of iterations. In the MRF algorithm, the number of classifications is set to 2 and the maximum number of iterations is 60. In the Otsu algorithm, the BlockSize is set to 3, and the average value is obtained by averaging. In the region growth algorithm, we use a 8-neighborhood growth setting.

\begin{figure*}
	\centering
	\includegraphics[scale=0.53]{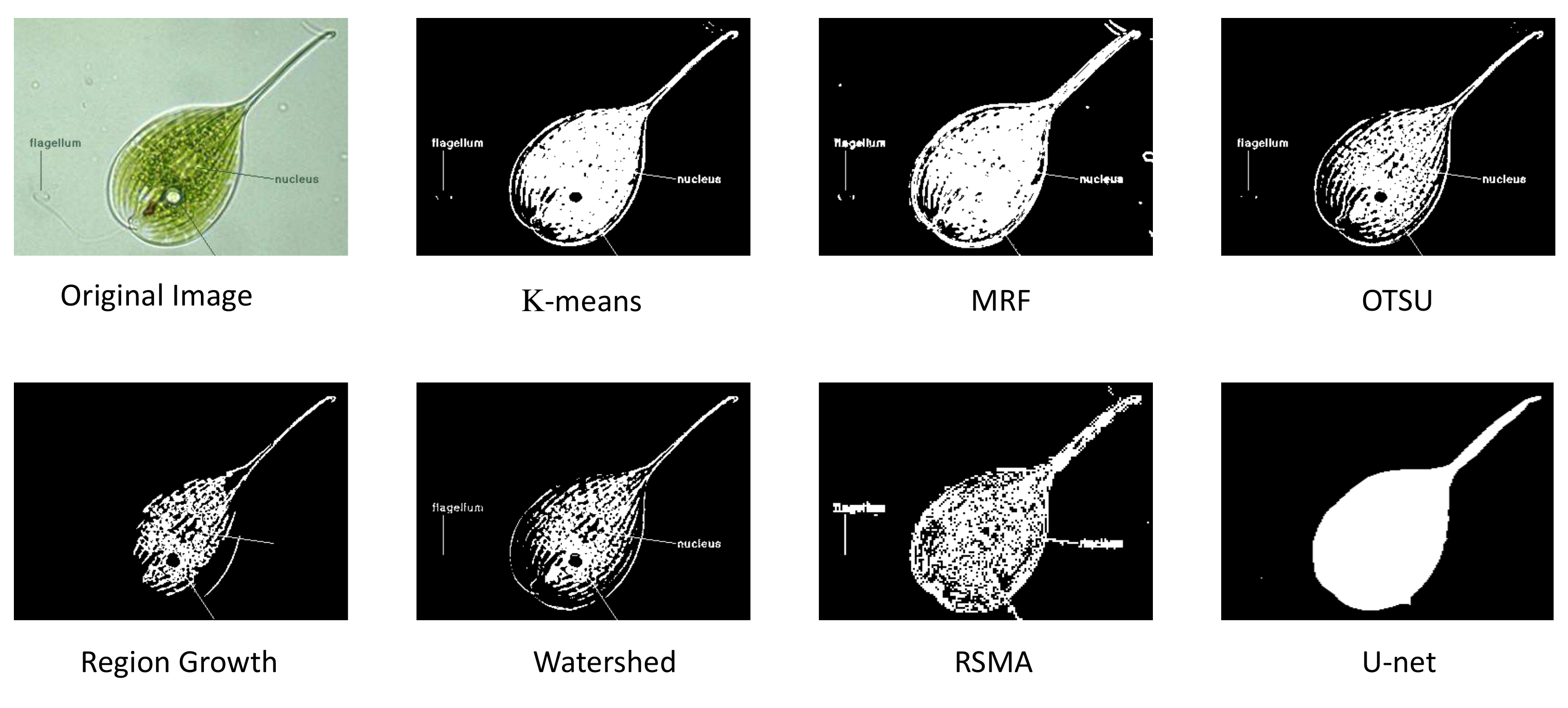}
	
	\caption{Output of results of different segmentation methods.}
	\label{FIG:3}
\end{figure*}

Among the seven classical segmentation methods, $k$-means is based on clustering, which is a region-based technology. Watershed algorithm is based on geomorphological analysis such as mountains and basins to implement different object segmentation algorithms. MRF is an image segmentation algorithm based on statistics. Its main features are fewer model parameters and strong spatial constraints. Otsu Thresholding is an algorithm based on global binarization, which can realize adaptive thresholds. The REG segmentation algorithm starts from a certain pixel and gradually adds neighboring pixels according to certain criteria. When certain conditions are met, the regional growth is terminated, and object extraction is achieved. The RSMA is first to determine a split and merge criterion. When splitting to the point of no further division, the areas with similar characteristics are integrated. Figure. \ref{FIG:3} shows a sample of the results of different segmentation methods on EMDS-6.

\subsubsection{Image Feature Extraction Methods}

This paper uses ten methods for feature extraction, including two-colour features, One is HSV (Hue, Saturation and Value) feature~\cite{junhua_2012_research}, and the other is RGB (Red, Green and Blue) colour histogram feature~\cite{kavitha_2016_texture}. The three texture features include the Local Binary Pattern (LBP)~\cite{ojala_2002_multiresolution}, the Histogram of Oriented Gradient (HOG)~\cite{dalal_2005_histograms} and the Gray-level Cooccurrence Matrix (GLCM)~\cite{qunqun_2013_extraction} formed by the recurrence of pixel gray Matrix. The four geometric features (Geo)~\cite{ramesha_2010_feature} include perimeter, area, long-axis and short-axis and seven invariant moment features (Hu)~\cite{hu_1962_visual}. The perimeter, area, long-axis and short-axis features are extracted from the GT image, while the rest are extracted from the original image. Finally, we use a support vector machine (SVM) to classify the extracted features. The classifier parameters are shown in Table. \ref{tbl6}.

\begin{table}[]
\caption{Parameter setting of EMDS-6 feature classification using SVM. C (penalty coefficient), decision function shape (DFS), The error value of stopping training (tol), Geometric features (Geo).}
\renewcommand\arraystretch{1.5} 
\begin{tabular}{cccccc}

\toprule
Feature & Kernel & C       & DFS & tol  & Max iter \\
\toprule
LBP     & rbf    & 50000   & ovr & 1e-3 & -1        \\
GLCM    & rbf    & 10000   & ovr & 1e-3 & -1        \\
HOG     & rbf    & 1000    & ovr & 1e-3 & -1        \\
HSV     & rbf    & 100     & ovr & 1e-3 & -1        \\
Geo    & rbf    & 2000000 & ovr & 1e-3 & -1        \\
Hu      & rbf    & 100000  & ovr & 1e-3 & -1        \\
RGB     & rbf    & 20      & ovr & 1e-3 & -1        \\
\toprule
\end{tabular}
\label{tbl6}
\end{table}

\subsubsection{Image Classification Methods \label{sec: Image Classification Methods}}
In this paper, we design the following two experiments to test whether the EMDS-6 dataset can compare the performance of different classifiers. Experiment 1: use traditional machine learning methods to classify images. This chapter uses Geo features to verify the classifier’s performance. Moreover, traditional classifiers used for testing includes, three $k$-Nearest Neighbor ($k$NN) classifiers ($k$=1, 5, 10)~\cite{abeywickrama_2016_k}], three Random Forests (RF) (tree=10, 20, 30)~\cite{ho_1995_random} and four SVMs (kernel function=rbf, polynomial, sigmoid, linear)~\cite{chandra_2018_survey}. The SVM parameters are set as follows: penalty parameter C = 1.0, the maximum number of iterations is unlimited, the size of the error value for stopping training is 0.001, and the rest of the parameters are default values.

\begin{table}[!htbp]
	\small
	\caption{Deep learning model parameters.}\label{tbl10}
	\renewcommand\arraystretch{2.2}
\begin{tabular}{p{3.7cm}p{3.7cm}}
		\toprule
		Parameter  & Parameter \\
		\midrule
		Batch Size , 32 &Epoch      , 100   \\
		
		Learning   , 0.002 & Optimizer  , Adam\\
		\bottomrule
	\end{tabular}
\end{table}

\begin{table}[!htbp]
	\small
	\caption{Evaluation metrics of segmentation method. $TP$ (True Positive), $FN$ (False Negative), $V_{pred}$ (Represents the foreground predicted by the model), $V_{gt}$ (represents the foreground in a GT image.).}\label{tbl4}
	\renewcommand\arraystretch{2.2}
	\begin{tabular}{p{3cm}p{3.5cm}}
		\toprule
		Indicators & Formula \\
		\midrule
		Dice  &  $\frac{2~\times~\lvert V_{pred}~\cap~ V_{gt} \rvert}{\lvert V_{pred}~\rvert~+~\lvert V_{gt}~\rvert}$ \\
		
		Jaccard  & $\frac{\lvert V_{pred}~\cap~ V_{gt} \rvert}{\lvert V_{pred}~\cup~ V_{gt} \rvert}$ \\
		
		Recall & $ \frac{TP}{TP~+~FN}$ \\

		\bottomrule
	\end{tabular}
\end{table}

\begin{table}[]
\caption{Classifier classification performance evaluation index. }\label{Formula}
\renewcommand\arraystretch{1.8}
\begin{tabular*}{\tblwidth}{@{} LL@{} }
\toprule
\makecell[c]{Evaluation Indicators} & \makecell[c]{Formula} \\
\midrule
\makecell[c]{Accuracy} & \makecell[c]{$\rm \frac{TP+TN}{TP + TN + FP + FN}$} \\
\makecell[c]{Precision}  & \makecell[c]{$\rm \frac{TP}{TP + FP}$} \\
\makecell[c]{F1-score} & \makecell[c]{$ 2 \times \frac{P \times R}{P + R}$} \\
\makecell[c]{Recall} & \makecell[c]{$\rm \frac{TP}{TP + FN}$} \\

\bottomrule
\end{tabular*}
\end{table}

\begin{table*}[htbp]
\caption{Similarity comparison between denoised image and original image. Types of noise (ToN), Denoising method (DM). (In [\%].). }
\renewcommand\arraystretch{1.5} 
\begin{tabular}{@{}ccccccccccc@{}}

\toprule
\label{tbl2}

ToN / DM          & TROF  & MF: (3×3) & MF: (5×5) & WF: (3×3) & WF: (5×5) & MaxF  & MinF  & GMF   & AMF   \\
 \midrule
PN     & 98.36 & 98.24   & 98.00   & 98.32   & 98.15   & 91.97  & 99.73 & 99.21 & 98.11 \\
MN:0.2       & 99.02 & 90.29   & 89.45   & 91.98   & 91.08   & 71.15 & 99.02 & 98.89 & 90.65 \\
MN:0.04       & 99.51 & 99.51   & 99.51   & 95.57   & 95.06   & 82.35 & 99.51 & 98.78 & 94.92 \\
GN 0.01-0   & 96.79 & 96.45   & 96.13   & 96.75   & 96.40   & 85.01 & 99.44 & 98.93 & 96.28 \\
GN 0.01-0.5 & 98.60 & 98.52   & 98.35   & 98.97   & 98.81   & 96.32 & 99.67 & 64.35 & 98.73 \\
GN 0.03-0   & 94.64 & 93.99   & 93.56   & 94.71   & 94.71   & 76.46 & 99.05 & 98.74 & 93.82 \\
GN 0.03-0.5 & 97.11 & 96.95   & 96.66   & 98.09   & 97.79   & 94.04 & 99.24 & 66.15 & 97.54 \\
SPN:0.01       & 99.28 & 99.38   & 99.14   & 99.60   & 99.37   & 95.66 & 99.71 & 99.44 & 99.16 \\
SPN:0.03       & 98.71 & 98.57   & 98.57   & 99.29   & 98.87   & 92.28 & 99.24 & 99.26 & 98.80 \\
PpN               & 98.45 & 98.53   & 98.30   & 99.46   & 99.02   & 96.30 & 99.04 & 99.61 & 98.61 \\
BGN               & 97.93 & 97.74   & 97.74   & 97.91   & 97.69   & 90.00 & 99.66 & 99.16 & 97.60 \\
PGN               & 96.97 & 96.63   & 96.33   & 97.16   & 96.85   & 85.82 & 99.47 & 98.98 & 96.47 \\
SN                & 97.90 & 97.97   & 97.75   & 99.27   & 98.63   & 99.27 & 98.63 & 99.64 & 98.15\\
	\bottomrule      
\end{tabular}
\end{table*}

\begin{table*}[htbp]
\renewcommand\arraystretch{1.5} 
\centering
	\caption{Comparison of variance between denoised image and original image. (In [\%].).}
\begin{tabular}{@{}ccccccccccc@{}}
\toprule
\label{tbl3}
ToN / DM          & \multicolumn{1}{c}{TROF} & \multicolumn{1}{c}{MF: (3×3)} & \multicolumn{1}{c}{MF: (5×5)} & \multicolumn{1}{c}{WF: (3×3)} & \multicolumn{1}{c}{WF: (5×5)} & \multicolumn{1}{c}{MaxF} & \multicolumn{1}{c}{MinF} & \multicolumn{1}{c}{GMF} & \multicolumn{1}{c}{AMF} \\
 \midrule
PN                & 1.49                     & 0.77                        & 1.05                        & 0.52                        & 0.66                        & 3.68                     & 2.99                     & 0.41                    & 0.88                    \\
MN,v: 0.2         & 32.49                    & 14.94                       & 15.65                       & 9.33                        & 11.36                       & 39.22                    & 32.49                    & 4.32                    & 13.35                   \\
MN,v: 0.04        & 10.89                    & 10.89                       & 10.89                       & 2.99                        & 3.71                        & 14.41                    & 10.89                    & 0.98                    & 4.28                    \\
GN,m: 0,v: 0.01   & 3.81                     & 3.06                        & 3.44                        & 2.06                        & 2.62                        & 11.68                    & 7.36                     & 1.16                    & 3.00                    \\
GN,m: 0.5,v: 0.01 & 0.89                     & 0.36                        & 0.41                        & 0.21                        & 0.28                        & 0.99                     & 1.74                     & 61.93                   & 0.43                    \\
GN,m: 0,v: 0.03   & 8.60                     & 7.78                        & 8.34                        & 5.04                        & 5.04                        & 27.23                    & 16.55                    & 4.24                    & 7.33                    \\
GN,m: 0.5,v: 0.03 & 1.60                     & 1.08                        & 1.18                        & 0.55                        & 0.73                        & 2.39                     & 3.06                     & 56.17                   & 1.05                    \\
SPN,d: 0.01       & 1.92                     & 1.21                        & 1.46                        & 0.10                        & 0.30                        & 6.37                     & 2.90                     & 4.73                    & 1.25                    \\
SPN,d: 0.03       & 3.84                     & 3.39                        & 3.39                        & 0.33                        & 1.09                        & 14.64                    & 5.18                     & 13.02                   & 3.15                    \\
PpN               & 2.88                     & 2.18                        & 2.44                        & 0.17                        & 0.72                        & 3.72                     & 4.48                     & 16.84                   & 2.09                    \\
BGN               & 2.35                     & 1.63                        & 1.94                        & 1.09                        & 1.38                        & 6.67                     & 4.57                     & 0.84                    & 1.66                    \\
PGN               & 3.79                     & 3.04                        & 3.42                        & 1.67                        & 2.13                        & 11.56                    & 7.33                     & 1.23                    & 2.98                    \\
SN                & 3.86                     & 3.17                        & 3.44                        & 0.31                        & 1.35                        & 4.82                     & 6.25                     & 5.58                    & 2.94                   \\   
	\bottomrule     
\end{tabular}
\end{table*}

\begin{table}[]
\caption{Evaluation of Feature Extraction methods using EMDS-6 dataset. (In [\%].).}
\renewcommand\arraystretch{1.5} 
\centering
\label{tbl5}
\begin{tabular}{p{2.5cm}p{1.2cm}p{1.2cm}p{1.2cm}}
\toprule

Method/Index                 & Dice & Jaccard & Recal \\
\toprule
$k$-means                      & 47.78                    & 31.38                       & 32.11                     \\
MRF                          & 56.23                    & 44.43                       & 69.94                     \\
Otsu                         & 45.23                    & 33.82                       & 40.60                     \\
REG                   & 29.72                    & 21.17                       & 26.94                     \\
RSMA & 37.35                    & 26.38                       & 30.18                     \\
Watershed                    & 44.21                    & 32.44                       & 40.75                     \\
U-Net                        & 88.35                    & 81.09                       & 89.67                  \\ 

\toprule        
\end{tabular}
\end{table}

\begin{table}[]

\caption{Different results obtained by applying different features in the EMDS-6 classification experiments using SVM. Feature type (FT),  Accuracy (Acc). (In [\%].).}
\renewcommand\arraystretch{1.5} 
\begin{tabular}{p{1.6cm}p{1.6cm}p{1.6cm}p{1.6cm}}

\toprule
FT    & LBP   & GLCM  & HOG   \\
\toprule
Acc   & 32.38 & 10.24 & 22.98 \\
HSV   & Geo   & Hu    & RGB   \\
29.52 & 50.0  & 7.86  & 28.81 \\
\toprule
\end{tabular}
\label{tbl7}
\end{table}

In Experiment 2, we use deep learning-based methods to classify images. Meanwhile, 21 classifiers are used to evaluate the performance, including, ResNet-18, ResNet-34, ResNet-50, ResNet-101~\cite{he_2016_ResNet}, VGG-11, VGG-13, VGG-16, VGG-19~\cite{simonyan_2014_VGG}, DenseNet-121, DenseNet-169~\cite{huang_2017_DenseNet}, Inception-V3~\cite{szegedy_2016_Inception-V3}, Xception~\cite{chollet_2017_xception}, AlexNet~\cite{krizhevsky_2012_AlexNet}, GoogleNet~\cite{szegedy_2015_GoogleNet}, MobileNet-V2~\cite{sandler_2018_mobilenetv2}, ShuffeleNetV2~\cite{ma_2018_shufflenet}, Inception-ResNet -V1~\cite{szegedy_2017_inception-ResNet-v2}, and a series of VTs, such as ViT~\cite{dosovitskiy_2020_ViT}, BotNet~\cite{srinivas_2021_BotNet}, DeiT~\cite{touvron_2020_Deit}, T2T-ViT~\cite{yuan_2021_T2T-ViT}. The above models are set with uniform hyperparameters, as detailed in Table. \ref{tbl10}.

\begin{table*}[htbp]

\caption{Results of experiments to classify Geo features using traditional classifiers. (In [\%].).}
\renewcommand\arraystretch{1.5} 
\begin{tabular}{p{2.8cm}p{2.2cm}p{2.7cm}p{2.2cm}p{2.4cm}p{2.4cm}}
\toprule

Classifier type & SVM: linear & SVM: polynomial & SVM: RBF  & SVM: sigmoid & RF,nT: 30 \\
\toprule
Accuracy              & 51.67       & 27.86           & 28.81     & 14.29        & 98.33     \\
$k$NN,k: 1       & $k$NN,k: 5   & $k$NN,k: 10      & RF,nT: 10 & RF,nT: 20    & --        \\
23.1            & 17.86       & 17.38           & 96.19     & 97.86        & --   \\
\toprule    
\end{tabular}
\label{tbl8}
\end{table*}

\begin{table*}
\caption{Classification results of different deep learning models. Accuracy (Acc), Params Size (PS)}
\renewcommand\arraystretch{1.2} 
\setlength{\tabcolsep}{6mm}{
\centering
\begin{tabular}{@{}lllllll@{} }
\toprule

Model             & Precision & Recall  & F1-Score & Acc &PS (MB) & Time (S)    \\
\toprule
Xception          & 44.29\%   & 45.36\% & 42.40\%  & 44.29\%  & 79.8        & 1079 \\
ResNet34          & 40.00\%   & 43.29\% & 39.43\%  & 40.00\%  & 81.3        & 862  \\
Googlenet         & 37.62\%   & 40.93\% & 35.49\%  & 37.62\%  & 21.6        & 845  \\
Densenet121       & 35.71\%   & 46.09\% & 36.22\%  & 35.71\%  & 27.1        & 1002 \\
Densenet169       & 40.00\%   & 40.04\% & 39.16\%  & 40.00\%  & 48.7        & 1060 \\
ResNet18          & 39.05\%   & 44.71\% & 39.94\%  & 39.05\%  & 42.7        & 822  \\
Inception-V3      & 35.24\%   & 37.41\% & 34.14\%  & 35.24\%  & 83.5        & 973   \\
Mobilenet-V2      & 33.33\%   & 38.43\% & 33.97\%  & 33.33\%  & 8.82        & 848 \\
InceptionResnetV1 & 35.71\%   & 38.75\% & 35.32\%  & 35.71\%  & 30.9        & 878  \\
Deit              & 36.19\%   & 41.36\% & 36.23\%  & 36.19\%  & 21.1        & 847  \\
ResNet50          & 35.71\%   & 38.58\% & 35.80\%  & 35.71\%  & 90.1        & 967  \\
ViT               & 32.86\%   & 37.66\% & 32.47\%  & 32.86\%  & 31.2        & 788  \\
ResNet101         & 35.71\%   & 38.98\% & 35.52\%  & 35.71\%  & 162         & 1101   \\
T2T-ViT           & 30.48\%   & 32.22\% & 29.57\%  & 30.48\%  & 15.5        & 863  \\
ShuffleNet-V2     & 23.33\%   & 24.65\% & 22.80\%  & 23.33\%  & 1.52        & 790   \\
AlexNet           & 32.86\%   & 34.72\% & 31.17\%  & 32.86\%  & 217         & 789   \\
VGG11             & 30.00\%   & 31.46\% & 29.18\%  & 30.00\%  & 491         & 958  \\
BotNet            & 28.57\%   & 31.23\% & 28.08\%  & 28.57\%  & 72.2        & 971  \\
VGG13             & 5.24\%    & 1.82\%  & 1.63\%   & 5.24\%   & 492         & 1023 \\
VGG16             & 4.76\%    & 0.23\%  & 0.44\%   & 4.76\%   & 512         & 1074 \\
VGG19             & 4.76\%    & 0.23\%  & 0.44\%   & 4.76\%   & 532         & 1119 \\

\bottomrule
\end{tabular}}
\label{tbl11}
\end{table*}

\subsubsection{Object Detection Method}
In this paper, we use Faster RCNN~\cite{ren_2015_faster} and Mask RCNN~\cite{he_2017_mask} to test the feasibility of the EMDS-6 dataset for object detection. Faster RCNN provide excellent performance in many areas of object detection. The Mask RCNN is optimized on the original framework of Faster RCNN. By using a better skeleton (ResNet combined with FPN) and the AlignPooling algorithm, Mask RCNN achieves better detection results than Faster RCNN.

In this experiment, the learning rate is 0.0001, the model Backbone is ResNet50, and the batch size is 2. In addition, we used 25\% of the EMDS-6 data as training, 25\% is for validation, and the rest is for testing.

\subsection{Evaluation Methods}

\subsubsection{Evaluation Method for Image Denoising}
This paper uses mean-variance and similarity indicators to evaluate filter performance. The similarity evaluation index can be expressed as~\ref{Eq(1)}, where $i$ represents the original image, $i_1$ represents the denoised image, $N$ represents the number of pixels, and $A$ represents the similarity between the denoised image and the original image. When the value of $A$ is closer to $1$, the similarity between the original image and the denoised image is higher, and the denoising effect is significant.

\begin{center}
\begin{align}
A = 1 - \frac{\sum_{i=1}^n\lvert i_1 - i \rvert}{N \times 255}
\label{Eq(1)}
\end{align}
\end{center}

The variance evaluation index can be expressed as Eq. \ref{Eq(2)}, where $S$ denotes the mean-variance, $L_{(i, j)}$ represents the value corresponding to the coordinates of the original image $(i, j)$, and $B_{(i, j)}$ the value associated with the coordinates of the denoised image $(i, j)$. When the value of $S$ is closer to $0$, the higher the similarity between the original and denoised images, the better the denoising stability.

\begin{center}
\begin{align}
S = 1 - \frac{\sum_{i=1}^n(L_{(i,j)}-B_{(i,j)})^2}{\sum_{i=1}^nL_{(i,j)}^2}
\label{Eq(2)}
\end{align}
\end{center}

\subsubsection{Evaluation Method for Image Segmentation}
We use segmented images and GT images to calculate Dice, Jaccard and Recall evaluation indexes. Among the three evaluation metrics, the Dice coefficient is pixel-level, and the Dice coefficient takes a range of 0-1. The more close to 1, the better the structure of the model. The Jaccard coefficient is often used to compare the similarity between two samples. When the Jaccard coefficient is larger, the similarity between the samples is higher. The recall is a measure of coverage, mainly for the accuracy of positive sample prediction. The computational expressions of Dice, Jaccard, and Recall are shown in Table. \ref{tbl4}.

\subsubsection{Evaluation Index of Image Feature Extraction}
Image features can be used to distinguish image classes. However, the performance of features is limited by the feature extraction method. In this paper, we select ten classical feature extraction methods. Meanwhile, the classification accuracy of SVM is used to evaluate the feature performance. The higher the classification accuracy of SVM, the better the feature performance.

\subsubsection{Evaluation Method for Image Classification}
In Experiment 1 of \ref{sec: Image Classification Methods}, we use only the accuracy index to judge the performance of traditional machine learning classifiers.The higher the number of EMs that can be correctly classified, the better the performance of this classifier. In Experiment 2, the performance of deep learning models needs to be considered in several dimensions. In order to more accurately evaluate the performance of different deep learning models, we introduce new evaluation indicators. The evaluation indexes and the calculation method of the indexes are shown in Table. \ref{Formula}. In Table. \ref{Formula}, TP means the number of EMs classified as positive and also labeled as positive. TN means the number of EMs classified as negative and also labeled as negative. FP means the number of EMs classified as positive but labeled as negative. FN means the number of EMs classified as negative but labeled as positive.

\subsubsection{Evaluation Method for Object Detection}
In this paper, Average Precision (AP) and Mean Average Precision (mAP) are used to evaluate the object detection results. AP is a model evaluation index widely used in object detection. The higher the AP, the fewer detection errors. AP calculation method is shown in Eq \ref{Eq(3)} and Eq \ref{Eq(4)}.

\begin{center}
\begin{align}
AP = \sum_{n=1}^N(r_{n+1}~-~r_{n})Pinterp(r_{n+1}) 
\label{Eq(3)}
\end{align}
\end{center}

\begin{center}
\begin{align}
Pinterp(r_{n+1})~=~max_{\hat{r}=r_{n+1}}=P(\hat{r})
\label{Eq(4)}
\end{align}
\end{center}

Among them, $r_n$ represents the value of the nth recall, and $p(\hat{r})$ represents the value of precision when the recall is $\hat{r}$. 

\section{ Experimental Results and Analysis}

\subsection{Experimental Results Analysis of Image Denoising}
We calculate the filtering effect of different filters for different noises. Their similarity evaluation indexes are shown in Table. \ref{tbl2}. From Table. \ref{tbl2}, it is easy to see that the GMF has a poor filtering effect for GN 0.01-0.5. The TROF and the MF have better filtering effects for MN:0.04.

\begin{figure*}
	\centering
	\includegraphics[scale=0.55]{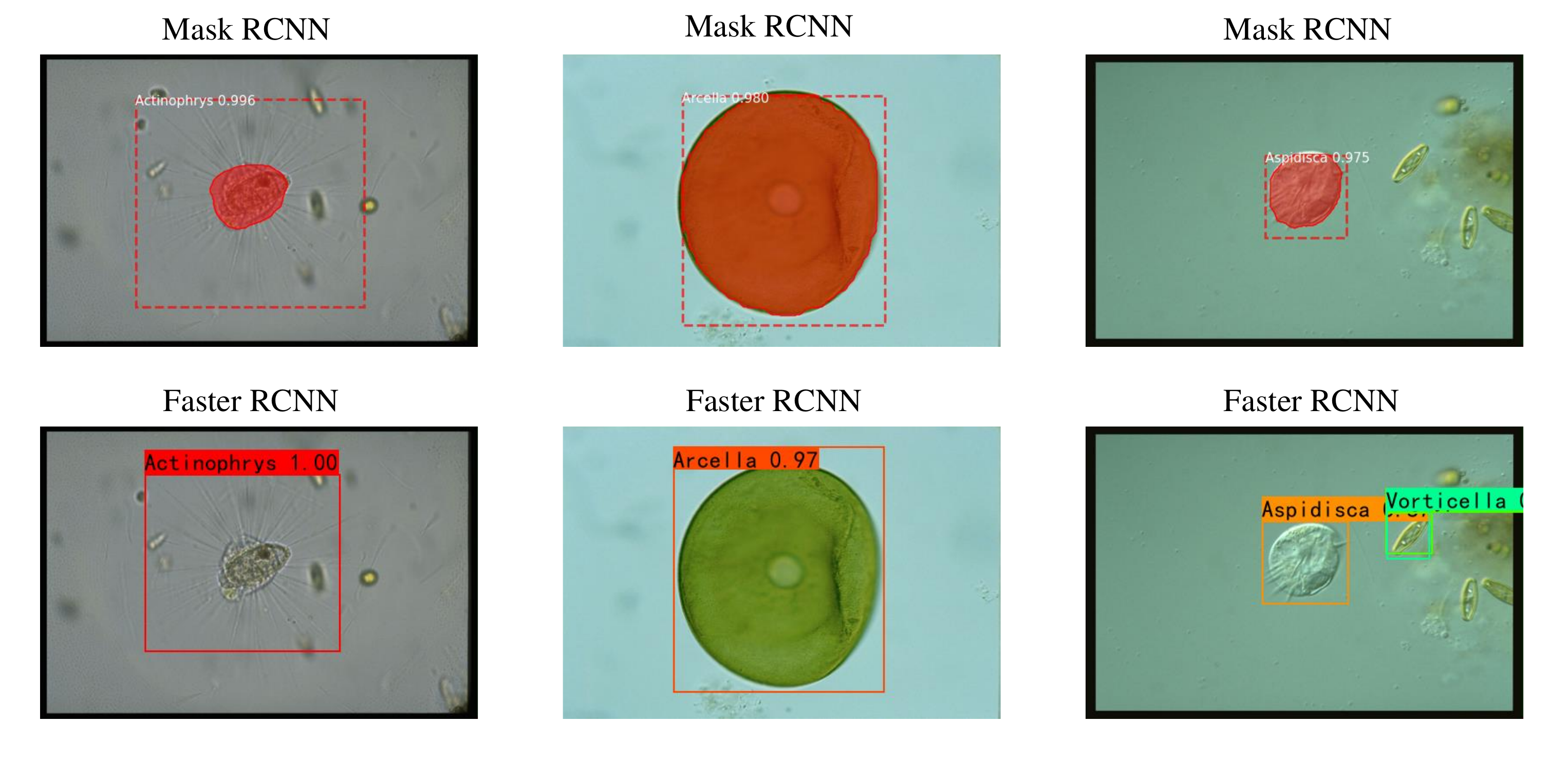}
	
	\caption{Faster RCNN and Mask RCNN object detection results.}
	\label{FIG:4}
\end{figure*}

\begin{table*}[]

\caption{AP and mAP based on EMDS-6 object detection of different types of EMs.}
\renewcommand\arraystretch{1.8}
\setlength\tabcolsep{3pt}%调列距
\begin{tabular}{ccccccccc}
\toprule
Model\textbackslash{}Sample (AP) & Actinophrys & Arcella    & Aspidisca & Codosiga    & Colpoda   & Epistylis   & Euglypha   & Paramecium \\
\toprule
Faster RCNN                 & 0.95        & 0.75       & 0.39      & 0.13        & 0.52      & 0.24        & 0.68       & 0.70       \\
Mask RCNN                   & 0.70         & 0.85       & 0.40       & 0.18       & 0.35      & 0.53       & 0.25       & 0.70        \\
Model\textbackslash{}Sample & Rotifera    & Vorticella & Noctiluca & Ceratium    & Stentor   & Siprostomum & K.Quadrala & Euglena    \\
Faster RCNN                 & 0.69        & 0.30       & 0.56      & 0.61        & 0.47      & 0.60        & 0.22       & 0.37       \\
Mask RCNN                   & 0.40         & 0.15       & 0.90      & 0.70         & 0.65      & 0.7         & 0.45       & 0.25       \\
Model\textbackslash{}Sample & Gymnodinium & Gonyaulax  & Phacus    & Stylongchia & Synchaeta & mAP         & --         & --         \\
Faster RCNN                 & 0.53        & 0.25       & 0.43      & 0.42        & 0.61      & 0.50      & --         & --         \\
Mask RCNN                   & 0.60         & 0.28      & 0.50       & 0.68       & 0.48     & 0.51      & --         & --       \\
\toprule 
\end{tabular}
\label{tbl12}
\end{table*}

\begin{table*}[]

\caption{EMDS history versions and latest versions. IC (Image Classification), IS (Image Segmentation), SoIS (Single-object Image Segmentation), MoIS (Multi-object Image Segmentation), SoFE (Single-object Feature Extraction), MoFE (Multi-object Feature Extraction), IR (Image Retrieval), IFE (Image Feature Extraction) IOD (Image Object Detection), IED (Image Edge Detection), ID (Image denoising), ECN (EM Class Number), OIN (Original Image Number), GTIN (Ground Truth Image Number), S (Single Object), M (Multiple object).}
\renewcommand\arraystretch{2.3}
\setlength\tabcolsep{3pt}%调列距
\begin{tabular}{llllcl}
\toprule 
Dataset                    & ECN & OIN & GTIN & Detaset Link                                                                                                          & Functions                                                                     \\ \toprule 
EMDS-1 \cite{li_2013_classification}    & 10  & 200 & 200  & - -                                                                                                                    & IC, IS                                                                        \\
EMDS-2 \cite{li_2013_classification}   & 10  & 200 & 200  & - -                                                                                                                    & IC ,IS                                                                        \\
EMDS-3 \cite{li_2016_environmental}      & 15  & 300 & 300  & - -                                                                                                                    & IC, IS                                                                        \\
EMDS-4 \cite{zou_2016_environmental}    & 21  & 420 & 420  & \multicolumn{1}{l}{\begin{tabular}[c]{@{}l@{}}\url{https://research.project-10.de/em-classiffication/}\end{tabular}}     & IC, IS, IR                                                                    \\
EMDS-5 \cite{li2021emds}  & 21  & 420 & \begin{tabular}[c]{@{}l@{}}840 \\ (S 420, M 420)\end{tabular}  & \multicolumn{1}{l}{\url{https://github.com/NEUZihan/EMDS-5}}                                                                & \begin{tabular}[c]{@{}l@{}}ID, IED, SoIS, MoIS,\\ SoFE, MoFE, IR\end{tabular} \\
EMDS-6 {[}In this paper{]} & 21  & 840 & 840  & \multicolumn{1}{l}{\begin{tabular}[c]{@{}l@{}}\url{https://figshare.com/articles/dataset/EMDS6/17125025/1}\end{tabular}} & ID, IC, IS, IFE, IOD  \\
\toprule 
\end{tabular}
\label{tbl13}
\end{table*}

In addition, the mean-variance is a common index to evaluate the stability of the denoising method. In this paper, the variance of the EMDS-6 denoised EM images and the original EM images are calculated as shown in Table. \ref{tbl3}. As the noise density increases, the variance significantly increases among the denoised and the original images. For example, by increasing the SPN density from 0.01 to 0.03, the variance increases significantly under different filters. This indicates that the result after denoising is not very stable.

From the above experiments, EMDS-6 can test and evaluate the performance of image denoising methods well. Therefore, EMDS-6 can provide strong data support for EM image denoising research.

\subsection{Experimental Result Analysis of Image Segmentation }

The experimental results of the seven different image segmentation methods are shown in Table. \ref{tbl5}. In Table. \ref{tbl5}, the REG and RSMA have poor segmentation performance, and their Dice, Jaccard, and Recall indexes are much lower than other segmentation methods. However, the deep learning-based, U-Net, has provided superior performance. By comparing these image segmentation methods, it can be concluded that EMDS-6 can provide strong data support for testing and assessing image segmentation methods.

\subsection{Experimental Result Analysis of Feature Extraction}

In this paper, we use the SVM to classify different features. The classification results are shown in Table. \ref{tbl7}. The Hu features performed poorly, while the Geo features performed the best. In addition, the classification accuracy of FT, LBP, GLCM, HOG, HSV and RGB features are also very different. By comparing these classification results, we can conclude that EMDS-6 can be used to evaluate image features.

\subsection{Experimental Result Analysis of Image Classification}
This paper shows the traditional machine learning classification results in Table. \ref{tbl8}, and the deep learning classification results are shown in Table~\ref{tbl11}. In Table. \ref{tbl8}, the RF classifier performs the best. However, the performance of the SVM classifier using the sigmoid kernel function is relatively poor. In addition, there is a big difference in Accuracy between other classical classifiers. From the computational results, the EMDS-6 dataset is able to provide data support for classifier performance evaluation.
According to Table.~\ref{tbl11},  the classification accuracy of Xception is 44.29\%, which is the highest among all models. The training of deep learning models usually consumes much time, but some models have a significant advantage in training time. Among the selected models, ViT consumes the shortest time in training samples. The training time of the ViT model is the least. The classification performance of the ShuffleNet-V2 network is average, but the number of parameters is the least. Therefore, experiments prove that EMDS-6 can be used for the performance evaluation of deep learning classifiers.

\subsection{Experimental Result Analysis of Image Object Detection}
The AP and mAP indicators for Faster CNN and Mast CNN are shown in Table.~\ref{tbl12}. We can see from Table.~\ref{tbl12} that Faster RCNN and Mask RCNN have very different object detection effects based on their AP value. Among them, the Faster RCNN model has the best effect on Actinophrys object detection. The Mask RCNN model has the best effect on Arcella object detection. Based on the mAP value, it is seen that Faster RCNN is better than Mask RCNN for object detection. The result of object detection is shown in Figure.~\ref{FIG:4}. Most of the EMs in the picture can be accurately marked. Therefore it is demonstrated that the EMDS-6 dataset can be effectively applied to image object detection.

\subsection{Discussion}
As shown in Table \ref{tbl13}, six versions of the EMs dataset are published. In the iteration of versions, different EMSs assume different functions. Both EMDS-1 and EMDS-2 have similar functions and can perform image classification and segmentation. In addition, both EMDS-1 and EMDS-2 contain ten classes of EMs, 20 images of each class, with GT images. Compared with the previous version, EMDS-3 does not add new functions. However, we expand five categories of EMs.

We open-source EMDSs from EMDS-4 to the latest version of EMDS-6. Compared to EMDS-3, EMDS-4 expands six additional classes of EMs and adds a new image retrieval function. In EMDS-5, 420 single object GT images and 420 multiple object GT images are prepared, respectively. Therefore EMDS-5 supports more functions as shown in Table \ref{tbl13}. The dataset in this paper is EMDS-6, which is the latest version in this series. EMDS-6 has a larger data volume compared to EMDS-5. EMDS-6 adds 420 original images and 420 multiple object GT images, which doubles the number of images in the dataset. With the support of more data volume, EMDS-6 can achieve more functions in a better and more stable way. For example, image classification, image segmentation, object and object detection.

\section{Conclusion and Future Work}
This paper develops an environmental microorganism image datasets, namely EMDS-6. EMDS-6 contains 21 types of EMs and a total of 1680 images. Including 840 original images and 840 GT images of the same size. Each type of EMs has 40 original images and 40 GT images. In the test, 13 kinds of noises such as multiplicative noise and salt and pepper noise are used, and nine kinds of filters such as Wiener filter and geometric mean filter are used to test the denoising effect of various noises. The experimental results prove that EMDS-6 has the function of testing the filter denoising effect. In addition, this paper uses 6 traditional segmentation algorithms such as $k$-means and MRF and one deep learning algorithm to compare the performance of the segmentation algorithm. The experimental results prove that EMDS-6 can effectively test the image segmentation effect. At the same time, in the image feature extraction and evaluation experiment, this article uses 10 features such as HSV and RGB extracted from EMDS-6. Meanwhile, the SVM classifier is used to test the features. It is found that the classification results of different features are significantly different, and EMDS-6 has the function of testing the pros and cons of features. In terms of image classification, this paper designs two experiments. The first experiment uses three classic machine learning methods to test the classification performance. The second experiment uses 21 deep learning models. At the same time, indicators such as accuracy and training time are calculated to verify the performance of the model from multiple dimensions. The results show that EMDS-6 can effectively test the image classification performance. In terms of object detection, this paper tests Faster RCNN and Mask RCNN, respectively. Most of the EMs in the experiment can be accurately marked. Therefore, EMDS-6 can be effectively applied to image object detection.

In the future, we will further expand the number of EM images of EMDS-6. At the same time, we will try to apply EMDS-6 to more computer vision processing fields to further promote microbial research development.

\section*{Acknowledgments}

This work is supported by the ``National Natural Science Foundation of China'' (No.61806047).  We thank Miss Zixian Li and Mr. Guoxian Li for their important discussion.

\section*{Declaration of Competing Interest}
The authors declare that they have no conflict of interest in this paper~\cite{2014k,chandra2021survey,kulwa2019state, zhang2021comprehensive,zhang2021lcu}.

\cite{li2015application, li2019survey, li2021state, rahaman2020identification}
\cite{srinivas2021bottleneck, mingqiang2008survey, zhao2022comparative}

\bibliographystyle{unsrt}
\bibliography{cas-refs}
\end{document}